\documentclass[letterpaper]{article}
\usepackage{aaai2027}
\usepackage[hyphens]{url}
\usepackage{graphicx}
\usepackage{natbib}
\usepackage{caption}
\usepackage{algorithm}
\usepackage{algorithmic}
\usepackage{booktabs}
\usepackage{amsmath}
\usepackage{amssymb}
\usepackage{placeins}

\nocopyright

\newcommand{\method}{\textsc{VCRT}}
\newcommand{\orbit}{\mathcal{O}}

\newcommand{\JS}{\mathrm{JS}}

\title{From Reasoning Strings to Partial Orders: Verifier-Certified Rule Transport through Quotient Policy Optimization}
\author{
Bang Xie\textsuperscript{\rm 1}, \quad Hao Liu\textsuperscript{\rm 1}, \quad
Zhiyuan Peng\textsuperscript{\rm 1}, \quad Xin Yin\textsuperscript{\rm 2},\\
Chenhao Ying\textsuperscript{\rm 1}\corresponding, \quad Yuan Luo\textsuperscript{\rm 1}, \quad
Senjian Zhang\textsuperscript{\rm 1}, \quad Wei Chen\textsuperscript{\rm 1}
}
\affiliations{
\textsuperscript{\rm 1}Shanghai Jiao Tong University, Shanghai, China\\
\textsuperscript{\rm 2}Zhejiang University, Hangzhou, China\\
yingchenhao@sjtu.edu.cn
}

\begin{document}
\maketitle

\begin{abstract}
Many computations admit several valid execution orders because independent subgoals or disjoint state updates can commute.
Reinforcement learning with verifiable rewards usually treats each successful trace as a separate token sequence, so serialization choices can be mistaken for logical dependencies.
We introduce Verifier-Certified Rule Transport (\method), which replays adjacent operation pairs with native verifiers.
Pairs whose two orders are accepted and reach the same canonical state provide commutation certificates; rejected or state-changing reversals provide anti-diamonds.
\method{} uses anti-diamonds to preserve genuine prerequisites and assigns policy credit to the total probability mass of each certified orbit.
It also constrains post-swap consistency, source retention, and policy drift.
We evaluate leave-one-environment-out transfer across ProofWriter, CLRS, and Lean through a shared anonymized relation-graph interface.
All training and checkpoint decisions are frozen before held-out evaluation, which uses one greedy trajectory per item without search or verifier feedback.
\method{} obtains a 77.60\% macro pass rate versus 64.53\% for the strongest matched baseline, a paired gain of 13.06 points (95\% bootstrap CI [12.58, 13.54]).
Lean accounts for most of this gain at 33.49 points, while ProofWriter and CLRS improve by 2.85 points on average.
Mechanism tests consistently favor anti-diamond supervision, whereas No-Orbit is statistically indistinguishable from full \method{}.
The evidence does not establish a general benefit from exact orbit aggregation.
\end{abstract}

\section{Introduction}

Autoregressive generation imposes an order on output tokens.
The computation represented by those tokens may allow several orders: proofs contain independent subgoals, algorithms update disjoint state components, and deductive systems apply unrelated rules in either sequence.
A generated trace is then one linear extension of a partially ordered computation.
Sequence-level supervision can reward that particular presentation instead of the dependencies that make it valid.

Reinforcement learning with verifiable rewards (RLVR) reduces ambiguity about final correctness by replacing a learned preference score with an executable check \cite{cobbe2021verifiers,shao2024deepseekmath,deepseekai2025r1}.
Group Relative Policy Optimization (GRPO) and related objectives can improve reasoning without a separately trained critic \cite{shao2024deepseekmath,mroueh2025grpo}.
Their optimization unit is still a sampled trajectory, which leaves open when two successful rollouts should share credit.
Two legal traces may differ only by commuting steps, yet trajectory-level optimization treats them as separate events.
Indiscriminate order augmentation creates the opposite error by rewarding invalid permutations.

Transfer requires a policy to distinguish harmless reordering from violated prerequisites.
A sequence model may overfit a familiar proof script, while indiscriminate order invariance can admit invalid reversals.
We therefore supervise relations between operations: commutation, precedence, and conflict.
Positive pairs share a verifier-certified outcome; negative pairs introduce one controlled ordering violation.

We ask whether native verifiers can provide the missing structural supervision.
ProofWriter supplies a deductive rule engine \cite{tafjord2021proofwriter}, CLRS exposes algorithmic state transitions \cite{velickovic2022clrs}, and Lean checks proof programs with a trusted kernel \cite{demoura2021lean4,yang2023leandojo}.
For a pair of operations, we replay both orders.
When both paths are accepted and reach the same canonical terminal state, they form a \emph{reasoning diamond}.
When reversal is rejected or changes the verified terminal state, the pair forms an \emph{anti-diamond}.
These local certificates distinguish commutativity from precedence without relying on textual similarity.

\begin{figure*}[t]
    \centering
    \includegraphics[width=0.94\textwidth]{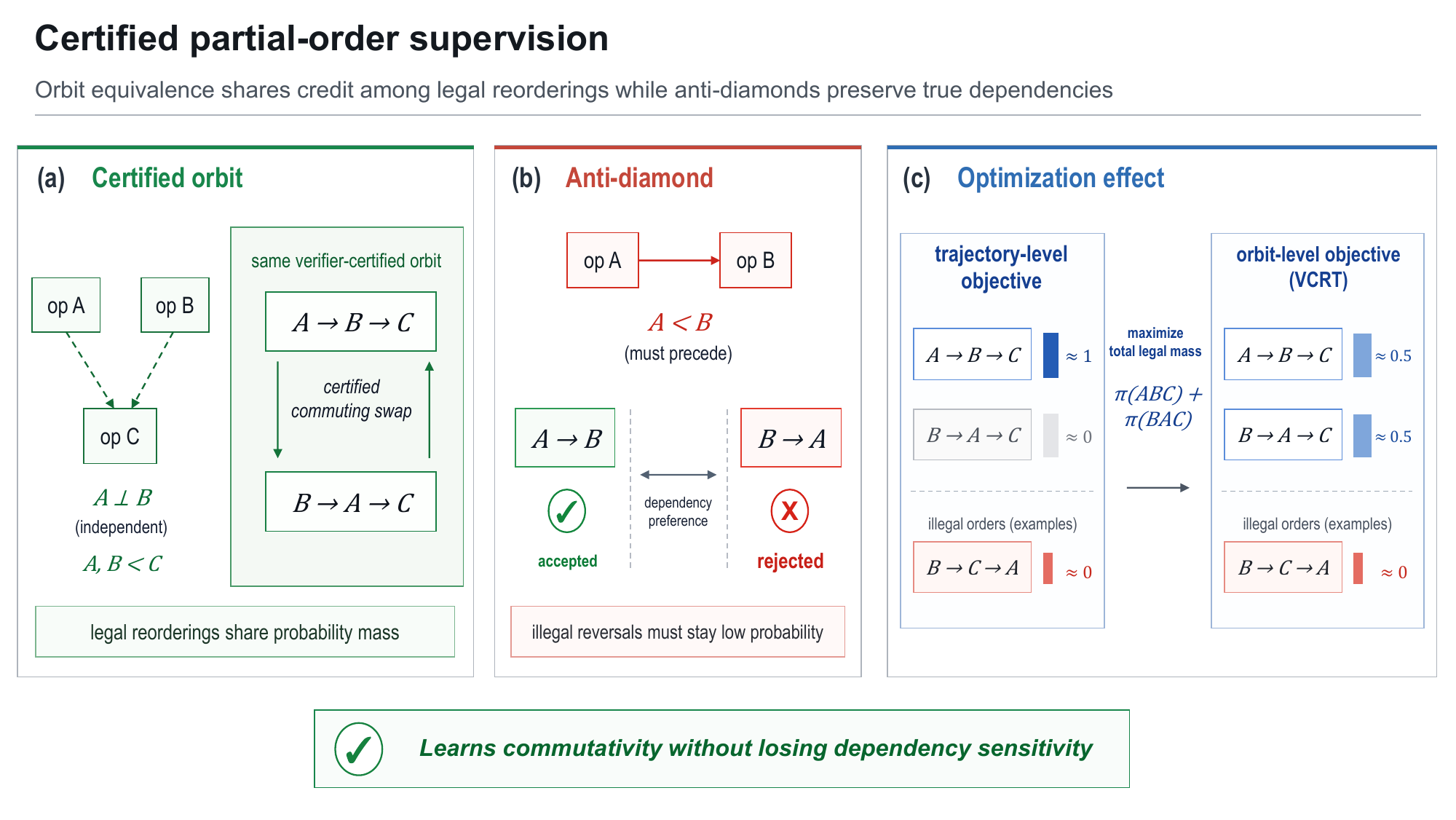}
    \caption{Verifier-certified partial-order supervision. \method{} shares credit across certified legal reorderings while separating dependency-violating reversals.}
    \label{fig:supervision}
\end{figure*}

Verifier-Certified Rule Transport (\method) converts these certificates into a quotient policy objective.
Successful linearizations connected by certified swaps form a verifier orbit.
\method{} applies the PPO ratio to the orbit's total policy mass.
An anti-diamond margin separates legal and illegal mass, while auxiliary constraints control post-swap action distributions, source retention, and action-level KL.
A dual-view interface keeps native states, verdicts, versions, and provenance in a sealed certificate view, while the policy receives only an anonymized relation graph.
Figure~\ref{fig:supervision} summarizes the resulting supervision.

Our evaluation holds out one complete reasoning environment at a time.
The model trains on the other two and is evaluated once on the third after every method, run configuration, and checkpoint has been frozen.
This setup measures whether a policy can use the same certified relation vocabulary in an unseen environment.
Dependency extraction from unrestricted text lies outside the experiment.
We contribute:
\begin{itemize}
    \item Native replay certificates for commuting operations and prerequisite-violating reversals. No-Anti underperforms full \method{} in each held-out fold.
    \item A quotient-policy objective over certified trajectory orbits, together with commutation, retention, and policy-drift constraints. The empirical benefit of exact orbit aggregation remains environment-dependent.
    \item A frozen leave-one-environment-out evaluation under a shared relation interface. \method{} gains 13.06 points over the strongest matched baseline, replicates positively on Granite-4.1-3B, and keeps source decline below one point.
\end{itemize}

\section{Related Work}

\paragraph{Reasoning traces, search, and tools.}
Chain-of-thought, self-consistency, STaR, and Minerva exploit intermediate or self-generated solutions, while search and tool-assisted methods broaden the executable trace set \cite{wei2022cot,wang2023selfconsistency,zelikman2022star,lewkowycz2022minerva,yao2023tot,gao2023pal,chen2023pot,yao2023react,schick2023toolformer}.
Such methods are commonly studied on verifiable benchmarks including MATH \cite{hendrycks2021math}.
For \method{}, multiple traces define a training-time equivalence relation only after verifier certification; inference uses one greedy trace.

\paragraph{Reinforcement learning with verifiable rewards.}
RLHF established preference-based policy learning \cite{christiano2017preferences,ouyang2022instructgpt}; PPO, DPO, and REINFORCE-style updates provide different policy controls \cite{schulman2017ppo,rafailov2023dpo,ahmadian2024reinforce}.
GRPO supports verifier-driven reasoning without a learned critic, while outcome and process verifiers scale supervision \cite{shao2024deepseekmath,mroueh2025grpo,cobbe2021verifiers,uesato2022process,lightman2024verify,deepseekai2025r1}.
\method{} retains the verifier reward but replaces the sampled-trajectory ratio with a certified-orbit ratio.

\paragraph{Policy equivalence and partial-order semantics.}
Trace semantics and partial-order reduction identify or avoid redundant independent-event interleavings \cite{mazurkiewicz1987trace,godefroid1996partialorder}.
MDP equivalence, abstraction, and homomorphisms formalize behavior-preserving state or policy quotients \cite{givan2003equivalence,li2006abstraction,vanderpol2020homomorphic}; POMO exploits symmetric optima in routing \cite{kwon2020pomo}.
Reasoning equivalence is instance-dependent.
\method{} certifies each local swap and terminal state before adding trajectories to an orbit.

\paragraph{Cross-domain and contrastive reasoning optimization.}
Recent multi-domain contrastive policy optimization aligns correct trajectories across domains and contrasts them with incorrect rollouts \cite{yu2026mcpo}, making it the closest policy-optimization precedent to our setting.
\method{} uses a different supervision object: positives are linearizations connected by verifier-certified swaps, negatives are matched dependency violations, and the PPO ratio is applied to certified orbit mass.
Our novelty claim is limited to this verifier-grounded equivalence construction and quotient ratio, not cross-domain contrastive learning itself.

\paragraph{Executable reasoning environments.}
Formal theorem proving combines generative models, curriculum learning, mathematical pretraining, benchmarks, and large-scale Lean proving \cite{polu2020gptf,polu2023curriculum,azerbayev2024llemma,zheng2022minif2f,xin2024deepseekprover}.
Verifier-guided logic systems use deduction, symbolic solvers, or machine-checkable theories \cite{ling2023deductive,pan2023logiclm,tafjord2021proofwriter}; CLRS and LeanDojo expose executable algorithmic and proof states \cite{velickovic2022clrs,yang2023leandojo,demoura2021lean4}.
We use these systems as certifiers of local state transitions: environment-private states establish supervision, while the learned policy receives only a common anonymous relation interface.

\section{Problem Formulation}

Let environment $e\in\mathcal{E}$ have states $\mathcal{S}_e$, actions $\mathcal{A}_e$, transition function $T_e$, and native verifier $V_e$.
A transition is certified when
\begin{equation}
 s \xrightarrow{a}_e s',\qquad V_e(s,a,s')=1.
\end{equation}
A trajectory $\tau=(a_1,\ldots,a_T)$ has policy probability
$\pi_\theta(\tau\mid x)=\prod_t\pi_\theta(a_t\mid h_t,x)$.

\subsection{Partial Orders and Linearizations}

For an instance $x$, let $P_x=(A_x,\prec_x)$ denote a finite strict partial order over its operations.
A complete trajectory is a \emph{linear extension} when it contains every operation once and respects every relation $a\prec_x b$.
The autoregressive model necessarily emits one linear extension, but the underlying partial order may admit many.
Let $\mathrm{LE}(P_x)$ be the legal set.
If two adjacent operations are incomparable under $\prec_x$, swapping them preserves membership in $\mathrm{LE}(P_x)$; repeated adjacent swaps connect all linear extensions of the same finite partial order.
In practice, however, $\prec_x$ is not given to the learner.
We therefore use native execution to certify individual swap edges and define only the connected component supported by those certificates.

\subsection{Diamonds and Anti-Diamonds}

Consider two local orders,
\begin{equation}
s\xrightarrow{a}s_a\xrightarrow{b'}s_{ab},
\qquad
s\xrightarrow{b}s_b\xrightarrow{a'}s_{ba}.
\end{equation}
They form a reasoning diamond when all four transitions are accepted, the adapted actions preserve their canonical signatures, and a frozen canonicalizer $\phi_e$ satisfies $\phi_e(s_{ab})=\phi_e(s_{ba})$.
An anti-diamond is a matched pair for which reversal is rejected, violates a precondition, or reaches a non-equivalent verified state.
When exactly one complete order is accepted, it certifies a precedence relation.

Two successful trajectories are verifier-equivalent, $\tau\sim_V\tau'$, if adjacent certified swaps transform one into the other.
Their orbit is
\begin{equation}
[\tau]_V=\{\tau':\tau'\sim_V\tau\}.
\end{equation}
The relation is local and certificate-based.
Equal final rewards or answers are insufficient because every connecting swap must have replay evidence.

In Lean, two independent sibling subgoals form a reasoning diamond when both execution orders reach the same verified terminal state.
If an action depends on an unproved lemma, reversing the pair is rejected by the kernel and provides an anti-diamond certificate for that prerequisite.

\section{Verifier-Certified Rule Transport}

\begin{figure*}[t]
    \centering
    \includegraphics[width=0.94\textwidth]{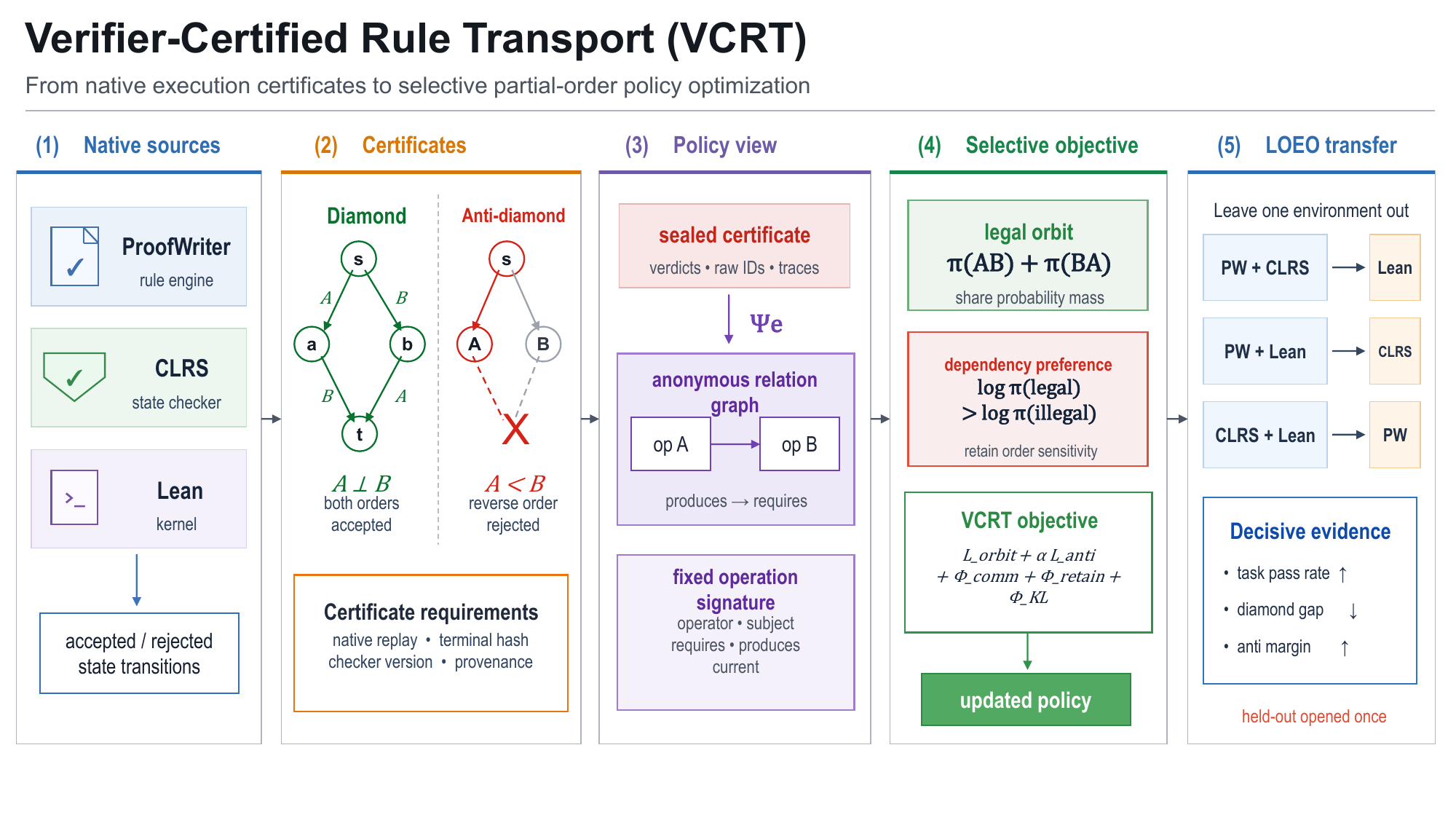}
    \caption{\method{} pipeline from native replay certificates to anonymized relation-graph training and leave-one-environment-out evaluation.}
    \label{fig:overview}
\end{figure*}

\subsection{Certificate and Policy Views}

Figure~\ref{fig:overview} shows the separation between certification and policy input.
The audit view retains native execution, verifier outcomes, canonical states, versioning, and provenance.
The policy view removes environment identifiers, verdict text, failure messages, and legal-action masks; it exposes only randomly renumbered action pointers and anonymous typed relations.
All methods receive this same policy view.

\subsection{Certificate Construction}

Construction begins from an accepted native transition and a candidate neighboring operation.
The adapter executes both local orders from the same pre-state, records all intermediate states, and checks every transition with the environment's native semantics.
A positive certificate requires four accepted transitions, matching operation signatures after adaptation, and identical frozen canonical hashes at the two terminal states.
For an anti-diamond, the legal order must be accepted and the minimally changed reversal must either be rejected or reach a distinct verified terminal hash.
Candidates with nondeterministic replay, ambiguous action correspondence, missing provenance, or checker-version disagreement are excluded rather than assigned a soft label.

After certification, the adapter writes the policy record while retaining enough audit information to reconstruct its label.

\subsection{Exact Orbit Mass}

For a verified orbit $\orbit$, define quotient-policy mass
\begin{equation}
\bar{\pi}_\theta(\orbit\mid x)
=Z_\theta(\orbit\mid x)
=\sum_{\tau\in\orbit}\pi_\theta(\tau\mid x).
\label{eq:orbitmass}
\end{equation}
Because the training instances are exactly enumerable, Equation~\ref{eq:orbitmass} uses the complete orbit sum.
Uncertified failures remain singleton orbits.

\paragraph{Representative invariance.}
Because Equation~\ref{eq:orbitmass} sums over the complete certified class, $Z_\theta([\tau]_V\mid x)=Z_\theta([\tau']_V\mid x)$ whenever $\tau\sim_V\tau'$.
The policy update therefore cannot depend on which representative of a successful orbit happened to be sampled.
A sampled subset would require a separate estimator and approximation analysis.

\paragraph{Reduction to ordinary trajectory optimization.}
When an orbit contains one trajectory, $Z_\theta(\{\tau\}\mid x)=\pi_\theta(\tau\mid x)$.
Thus failures and instances without certified commutations use the ordinary trajectory likelihood ratio.
\method{} changes credit assignment only where the verifier supplies evidence for equivalence.

\paragraph{Why optimize mass rather than uniformity.}
Averaging log likelihood over legal linearizations would favor a particular distribution inside the legal set and would change when additional equivalent traces are enumerated.
Orbit mass instead treats the certified set as one event.
Moving probability between two legal members leaves the target unchanged as long as their total mass is preserved.
Exact enumeration makes that quotient literal in the four-operation setting; illegal reversals remain outside the orbit and are handled by the separation loss.

\subsection{Orbit-GRPO}

At the start of an update, the behavior policy $\pi_k$, sampled trajectories, advantages, and orbit masses are frozen.
For group size $G$, the leave-one-out advantage is
\begin{equation}
A_i=R_i-\frac{1}{G-1}\sum_{j\neq i}R_j.
\end{equation}
For a successful sampled trajectory in orbit $\orbit_i$, the ratio is
\begin{equation}
\rho_i^{\orbit}(\theta)
=\frac{Z_\theta(\orbit_i\mid x_i)}
       {Z_k(\orbit_i\mid x_i)}.
\end{equation}
For singleton failures, it reduces to the ordinary trajectory ratio.
Below, $\rho_i$ denotes $\rho_i^{\orbit}$ for a certified successful orbit and the ordinary trajectory ratio for a singleton failure.
The policy loss is
\begin{equation}
\mathcal{L}_{\mathrm{orbit}}
=-\frac{1}{G}\sum_i
\min\!\left(\rho_iA_i,\,
\operatorname{clip}(\rho_i,1-\epsilon,1+\epsilon)A_i\right).
\label{eq:orbitgrpo}
\end{equation}
Applying the ratio to total legal mass gives alternatives credit in proportion to their current probability without forcing a uniform distribution over traces.
The gradient of $\log Z_\theta(\orbit\mid x)$ is a posterior-weighted average of trajectory score gradients:
\begin{equation}
\nabla_\theta\log Z_\theta(\orbit\mid x)
=\sum_{\tau\in\orbit}
\frac{\pi_\theta(\tau\mid x)}{Z_\theta(\orbit\mid x)}
\nabla_\theta\log\pi_\theta(\tau\mid x).
\label{eq:orbitgradient}
\end{equation}
Consequently, legal alternatives receive credit in proportion to their current mass without forcing a uniform distribution over stylistically different traces.

\subsection{Dependency Separation and Constraints}

For any finite trajectory set $S$, let $Z_\theta(S\mid x)=\sum_{\tau\in S}\pi_\theta(\tau\mid x)$.
Let $\orbit^+$ and $\orbit^-$ denote the legal and illegal trajectory sets for an anti-diamond.
Their log-mass margin is
\begin{equation}
\Delta_\theta=\log Z_\theta(\orbit^+\mid x)
              -\log Z_\theta(\orbit^-\mid x),
\end{equation}
and the separation loss is
\begin{equation}
\mathcal{L}_{\mathrm{anti}}
=\frac{1}{2}\max(0,m-\Delta_\theta)^2.
\end{equation}
For two certified swap histories $h_{ab}$ and $h_{ba}$, a canonical action alignment $M$ defines
\begin{equation}
\mathcal{L}_{\mathrm{comm}}
=\JS\!\left(M_{\#}\pi_\theta(\cdot\mid h_{ab})
\,\|\,M_{\#}\pi_\theta(\cdot\mid h_{ba})\right).
\end{equation}
The frozen map $M$ uses a bijection over canonical operation signatures; ambiguous pairs are excluded.
Batch constraints cap expected commutation loss at 0.01, source-orbit mass decline at 0.05, and action KL at 0.20.
The complete objective is
\begin{equation}
\mathcal{L}_{\method}
=\mathcal{L}_{\mathrm{orbit}}
+\alpha\mathcal{L}_{\mathrm{anti}}
+\Phi_{\mathrm{comm}}+\Phi_{\mathrm{retain}}+\Phi_{\mathrm{KL}}.
\label{eq:full}
\end{equation}
Here, each $\Phi$ is an augmented-Lagrangian penalty for the corresponding batch constraint.

\subsection{Optimization Semantics}

The anti-diamond loss and three auxiliary constraints have different roles.
The anti-diamond loss distinguishes prerequisites from legal swaps; the commutation constraint aligns next-action distributions after equivalent histories; retention bounds source decline; and action KL limits movement from the behavior policy.
The checker fixes orbit membership before optimization, so these terms shape probability allocation within that partition.

Each update freezes replay verdicts, advantages, and old-policy probabilities, then recomputes only new-policy quantities during optimization.

\paragraph{Training and inference boundary.}
Native replay is used to construct supervision on source episodes, not as a decoding tool.
During training, certificates determine which sampled successes share an orbit and which matched reversals receive the anti-diamond penalty.
At held-out inference, the model receives the anonymous relation record, emits one pointer sequence, and is scored once by the native checker.
The reported improvement therefore cannot be attributed to test-time search, repair, or verifier-guided resampling.

\begin{algorithm}[t]
\caption{\method{} Online Update}
\label{alg:vcrt}
\begin{algorithmic}[1]
\REQUIRE Source environments, native verifiers, policy $\pi_\theta$
\STATE Freeze $\pi_k$ and sample $G$ complete trajectories per prompt.
\STATE Replay trajectories and form exact certified orbits.
\STATE Freeze rewards, leave-one-out advantages, and old orbit masses.
\FOR{two optimizer epochs}
    \STATE Recompute new trajectory and orbit masses.
    \STATE Evaluate Equations~\ref{eq:orbitgrpo} through~\ref{eq:full}.
    \STATE Update LoRA parameters and constraint duals.
\ENDFOR
\end{algorithmic}
\end{algorithm}

\section{Experimental Design}

\paragraph{Research question.}
Does verifier-certified dependency supervision improve transfer to a held-out reasoning environment, and which components account for the gain over outcome GRPO and single-trajectory canonical GRPO?

\paragraph{Environments and isolation.}
The three environments are ProofWriter, CLRS, and Lean.
Each leave-one-environment-out (LOEO) fold trains on two environments and withholds the third.
Each environment contributes five frozen schemas with 500 episodes per schema.
Each episode contains four operations and 24 complete candidate orders, of which exactly 12 form the certified positive orbit.
Every fold therefore contains 5,000 source episodes and 120,000 replayed trajectories.
Native replay consistency is 100\%.
Held-out tasks, labels, reference traces, and verdicts are unavailable to training, model selection, and checkpoint selection.

\paragraph{Generalization target.}
The split withholds an entire native execution system rather than a random subset of episodes.
Action identifiers are renumbered and environment names are absent from the policy record, while the relation vocabulary is shared.
Success therefore requires reusing decisions associated with commutation, precedence, and conflict under new native semantics.
It does not require the model to recover those relations from raw source language, because certification and policy learning are deliberately separated.

\begin{table}[t]
\centering
\small
\caption{Frozen LOEO construction. Each source contributes five schemas with 500 episodes per schema. ``Replay'' counts all candidate orders checked during construction.}
\label{tab:data}
\begin{tabular}{@{}lccc@{}}
\toprule
Held out & Source environments & Episodes & Replay \\
\midrule
Lean & ProofWriter + CLRS & 5,000 & 120,000 \\
ProofWriter & CLRS + Lean & 5,000 & 120,000 \\
CLRS & ProofWriter + Lean & 5,000 & 120,000 \\
\bottomrule
\end{tabular}
\end{table}

\paragraph{Schema coverage and full-chain audit.}
Table~\ref{tab:data} reports the frozen construction.
Every fold must match these counts and achieve 100\% replay agreement before training.

\paragraph{Models and optimization.}
The primary backbone is Qwen3.5-2B at a hash-locked revision.
All methods use LoRA \cite{hu2022lora} with rank 64, scale 128, dropout 0.05, and attention and MLP projection targets.
Training runs for 400 updates, two source prompts per update, eight generations per prompt, and two optimizer epochs.
AdamW uses learning rate $5{\times}10^{-6}$; the paper checkpoint is update 400.

\paragraph{Methods.}
We compare \textsc{Outcome-GRPO}, which uses native terminal reward and a trajectory ratio; \textsc{Canonical-GRPO}, which shares the relational pointer interface but has no orbit objective; and \method{}, which uses Equations~\ref{eq:orbitmass} through~\ref{eq:full}.
Each core method runs three LOEO folds and three independent runs.
The mechanism study evaluates \textsc{No-Orbit}, which uses singleton ratios, and \textsc{Shuffled-Orbit}, which replaces the certified 12-of-24 orbit with a frozen random set of equal size, across all three folds and runs.
\textsc{No-Anti}, which removes the anti-diamond margin, is evaluated in a prespecified dedicated ablation.

\paragraph{Matched-comparison controls.}
All trained methods start from the same hash-locked backbone, see the same episode order, and use the same update count and checkpoint schedule.
Canonical-GRPO shares the anonymous relation interface with \method{}, controlling for the representation.
The component ablations distinguish effects within the bundled \method{} objective.

\paragraph{Evaluation.}
Held-out evaluation opens once, after all prespecified runs reach update 400 and their checkpoints are frozen.
Each item receives one temperature-zero greedy trajectory.
There is no search, repair, reranking, retrieval, self-consistency, or verifier feedback during generation.
The native checker scores the final trajectory.
Primary outcomes are held-out pass rate, source retention, three-fold direction, and a paired bootstrap 95\% confidence interval for \method{} against the strongest matched baseline.

\paragraph{Statistical analysis.}
Accuracy and verifier pass rate coincide because every generated complete pointer trajectory receives one deterministic native replay verdict.
We select the stronger of Outcome-GRPO and Canonical-GRPO by its prespecified three-fold macro average, then compute episode-paired differences and a 95\% bootstrap interval using a fixed random state.
Source retention is the difference between each trained checkpoint and its matched frozen backbone; the guardrail is a maximum decline of two absolute percentage points.

\paragraph{Unit of comparison.}
All methods are evaluated on the same held-out items, so the primary interval resamples paired item-level score differences rather than pooling unrelated trajectories.
Run means describe optimizer variation, fold means describe transfer direction, and the macro average gives each held-out environment equal weight.
These summaries are reported separately because a narrow pooled interval does not imply that every run or environment improves.

\section{Results}

\begin{table*}[t]
\centering
\caption{Held-out pass rate (\%) with one greedy trajectory per item. Each entry is the mean over three runs; ``Macro'' averages all three LOEO directions.}
\label{tab:main}
\begin{tabular}{lcccc}
\toprule
Method & Held-Out Lean & Held-Out ProofWriter & Held-Out CLRS & Macro \\
\midrule
Outcome-GRPO & 57.35 & 54.69 & 61.12 & 57.72 \\
Canonical-GRPO & 66.51 & 55.84 & 71.25 & 64.53 \\
\method{} & \textbf{100.00} & \textbf{57.95} & \textbf{74.84} & \textbf{77.60} \\
\bottomrule
\end{tabular}
\end{table*}

\begin{table*}[t]
\centering
\caption{Mechanism ablations. Entries are full \method{} minus the ablation in held-out pass-rate points, so positive values favor \method{}. Fold columns average three runs; intervals are shown where available.}
\label{tab:mechanism}
\small
\begin{tabular}{lrrrrr}
\toprule
Ablation & ProofWriter & CLRS & Lean & Macro & 95\% CI \\
\midrule
No-Anti & 8.48 & 4.68 & 23.12 & 12.09 & n/a \\
No-Orbit & $-0.08$ & $-1.65$ & 0.00 & $-0.58$ & [$-3.38$, 2.36] \\
Shuffled-Orbit & $-8.43$ & 5.40 & 31.19 & 9.39 & [$-7.71$, 32.17] \\
\bottomrule
\end{tabular}
\end{table*}

\subsection{Transfer under the Shared Interface}

Table~\ref{tab:main} reports the matched comparison.
Against Canonical-GRPO, \method{} gains 13.06 macro pass-rate points, with an episode-paired 95\% bootstrap CI of [12.58, 13.54].
The fold gains are 33.49 points on Lean, 2.11 on ProofWriter, and 3.59 on CLRS.
Lean contributes approximately 85\% of the macro gain; ProofWriter and CLRS improve by 2.85 points on average.
The held-out Lean score is 100\%, but five four-operation sibling-goal schemas may impose a ceiling on this fold.
Averaging by run also yields positive gains, although two of nine fold-by-run comparisons are negative.
The improvement is consistent in the fold and run aggregates, but not in every individual comparison.
Canonical-GRPO itself improves over Outcome-GRPO by 9.16 points on Lean, 1.15 on ProofWriter, and 10.13 on CLRS.
Thus the anonymous relational interface accounts for part of the gain over outcome-only training.
The additional increase from 64.53 to 77.60 macro pass rate is the quantity attributable to the complete \method{} objective under the matched interface.

\subsection{Mechanism and Ablations}

Table~\ref{tab:mechanism} shows that No-Anti is worse in every environment, while No-Orbit is statistically indistinguishable from full \method{} on average.
Shuffled-Orbit is worse on Lean and CLRS but better on ProofWriter, and its pooled interval crosses zero.
The prespecified ``all ablations decrease'' criterion is therefore not met.
Across the ablations, support is strongest for verified dependency separation; exact orbit aggregation remains environment-dependent.
The signs are informative.
Removing the anti-diamond margin reduces all three held-out means, including the two folds with modest main-comparison gains.
By contrast, replacing the orbit ratio with singleton ratios changes the macro result by less than one point, and shuffling the orbit changes both sign and magnitude across environments.
The component evidence therefore favors learning which orders are illegal more clearly than redistributing credit among all legal orders.

\subsection{Retention, Robustness, and Compute}

The largest source-task decline across the Qwen runs is 0.96 points, satisfying the prespecified two-point guardrail.
On Granite-4.1-3B, \method{} improves the frozen canonical backbone by 36.36 points on Lean, 9.96 on ProofWriter, and 6.40 on CLRS; the three-fold paired gain is 17.57 points (95\% CI [16.35, 18.83]).
Training consumes 376.55 aggregate GPU hours.
Both backbones improve in every held-out direction, so the positive macro result is not tied to one initialization.
The replication supports the direction of transfer, but it is not used to claim a scaling trend.

\begin{table}[t]
\centering
\small
\caption{Held-out gains over the matched canonical baseline. The Qwen row compares with Canonical-GRPO; the Granite row compares with its frozen canonical backbone.}
\label{tab:backbone}
\begin{tabular}{@{}lrrrr@{}}
\toprule
Backbone & Lean & ProofWriter & CLRS & Macro \\
\midrule
Qwen3.5-2B & 33.49 & 2.11 & 3.59 & 13.06 \\
Granite-4.1-3B & 36.36 & 9.96 & 6.40 & 17.57 \\
\bottomrule
\end{tabular}
\end{table}

\section{Discussion}
\label{sec:discussion}

The matched comparisons separate the input representation from part of the training objective, but they do not assign the full gain to a single component.
Outcome-GRPO uses terminal reward with a trajectory ratio, and Canonical-GRPO adds the same relational pointer interface used by \method{}.
Canonical-GRPO therefore controls for that interface.
The remaining 13.06-point macro gap belongs to the complete \method{} objective, which bundles orbit credit, anti-diamond separation, commutation consistency, retention, and KL control.

Anti-diamond supervision has the clearest component-level evidence.
Removing its margin hurts every held-out fold, even though the size of the loss varies by environment.
This result ties the gain to sensitivity to verified prerequisites within the bundled objective.
It does not isolate whether the margin acts independently of the commutation and retention terms, so we avoid attributing the entire macro improvement to anti-diamonds alone.

The orbit ablations tell a different story.
No-Orbit is 0.58 points above full \method{} on the macro average, with an interval that includes zero.
Shuffled-Orbit loses heavily on Lean, gains on ProofWriter, and also has an interval crossing zero.
Exact orbit mass may interact with the density and type of certified commutations, but the current folds do not show a stable average advantage.
The experiments support the semantic partition between legal and illegal orders more clearly than the choice to aggregate all legal orders into one ratio.

\paragraph{Search-free interpretation.}
The held-out scores come from one temperature-zero trajectory, so the comparison concerns the policy learned during source training.
There is no larger candidate budget for \method{} and no opportunity to ask the checker which branch to keep.
This matters because certified commutations enlarge the supervised success event during optimization, but they do not enlarge the inference budget.
The gain is therefore consistent with a policy that has learned to respect portable dependency relations, rather than a decoder that succeeds by exploring more orders.

Lean contributes about 85\% of the macro gain and reaches 100\% on five four-operation sibling-goal schemas.
ProofWriter and CLRS improve by 2.85 points on average, and two of nine fold-by-run comparisons are negative.
The present Lean setting cannot distinguish strong transfer from ceiling saturation.
The 0.96-point maximum source decline also makes broad source-task forgetting an unlikely account of the held-out gain.

Finally, the experiment measures transfer through a shared, pre-certified relation interface.
The policy receives anonymous pointers instead of environment names or raw verifier output, but it is still given the relations required for the decision.
The reported result is therefore about using a common relational vocabulary in a held-out environment.
Dependency extraction from unrestricted language and transfer across unrelated input modalities remain outside the evidence.

\paragraph{Engineering contract.}
Adding an environment does not require exposing its native state to the policy.
It requires an adapter that can replay two local orders from one pre-state, match the operations across those replays, and compare canonical terminal states under a fixed verifier version.
The resulting certificate can then be reduced to the same anonymous pointer-and-relation record used by the other environments.
This boundary is important operationally: environment-specific semantics stay in certificate construction, while training and evaluation consume one shared policy format.
The 100\% replay-agreement gate in Table~\ref{tab:data} checks that boundary before optimization begins.

\paragraph{What the results support.}
The experiments separate three claims that would otherwise be easy to conflate.
First, Canonical-GRPO shows that the shared relation representation is useful even without orbit optimization.
Second, the remaining matched gain shows that the bundled verifier-certified objective adds value beyond that representation.
Third, the ablations identify dependency separation as the most stable part of the bundle.
They do not provide comparable evidence that summing all certified legal orders into one ratio is necessary.
This narrower conclusion follows the observed folds and avoids using the overall gain as evidence for every term in the objective.

\section{Limitations and Broader Impact}

\paragraph{Verifier and representation scope.}
\method{} requires deterministic verifiers and a supplied certified relation graph.
The experiment therefore measures use of verified dependencies in a held-out environment rather than induction of those dependencies from raw text.

\paragraph{Scale and external validity.}
The matched result concerns compact backbones, three structured environments, and a shared greedy-evaluation interface.
Exact enumeration is practical for the four-operation episodes studied here; longer traces, larger models, and less formal environments remain outside the reported evidence.

\paragraph{Reproducibility essentials.}
Reproduction requires the frozen five-by-500 schema construction, all 24 candidate orders, update-400 checkpoints, matched item pairing, and fixed backbone, verifier, and canonicalizer versions.

\paragraph{Broader impact.}
A checker error can be reused across many updates, propagating its effect beyond one trajectory.
Deployments should therefore validate checkers independently and report certificate error rates.

\section{Conclusion}

Across the three LOEO folds, \method{} gains 13.06 points (CI [12.58, 13.54]) over the strongest matched baseline, although Lean contributes most of the improvement.
No-Anti is worse in every held-out fold, whereas replacing orbit aggregation with singleton ratios does not reduce the macro score.
The current evidence favors verifier-certified dependency separation as the transferable component and leaves exact orbit-mass optimization unresolved.

\clearpage
\bibliography{references}

\end{document}